%% file: f2fcss.tex
\documentclass[11pt,twoside,twocolumn,a4paper]{article}

\usepackage{acvrw}
\usepackage{times}
\usepackage{epsfig}
\usepackage{graphicx}
\usepackage{amsmath}
\usepackage{amssymb}

\makeatletter
\@namedef{ver@everyshi.sty}{}
\makeatother
\usepackage{tikz}
\definecolor{highlight}{RGB}{255,150,0}
\definecolor{highlight2}{RGB}{6,244,241}
\definecolor{crop}{RGB}{114,255,0}

\usepackage{units}
\usepackage{subfig}
\usepackage{makecell}
\usepackage{multirow}
\usepackage[font=small]{caption}
\usepackage{setspace}
\newcommand{\mat}[1]{\ensuremath{\mathbf{#1}}}
 \renewcommand{\vec}[1]{\ensuremath{\mathbf{#1}}}
\DeclareMathOperator{\prelu}{prelu}
\newcommand{\Cons}{Cons}
\newcommand{\ConsW}{ConsW}
\usepackage[pagebackref=true,breaklinks=true,letterpaper=true,colorlinks,bookmarks=false]{hyperref}

\acvrwfinalcopy  

\def\acvrwPaperID{10}

\ifacvrwfinal\fi
\begin{document}

\title{Frame-To-Frame Consistent Semantic Segmentation}

\author{Manuel Rebol \qquad Patrick Kn\"obelreiter\vspace{2pt}\\ 
Graz University of Technology\\
{\tt\small rebol@student.tugraz.at, knoebelreiter@icg.tugraz.at}
}

\maketitle
\ifacvrwfinal\thispagestyle{fancy}\fi

\begin{abstract} 
	In this work, we aim for temporally consistent semantic segmentation throughout frames in a video. Many semantic segmentation algorithms process images individually which leads to an inconsistent scene interpretation due to illumination changes, occlusions and other variations over time. 
	To achieve a temporally consistent prediction, we train a convolutional neural network (CNN) which propagates features through consecutive frames in a video using a convolutional long short term memory (ConvLSTM) cell. 
	Besides the temporal feature propagation, we penalize inconsistencies in our loss function. 
	We show in our experiments that the performance improves when utilizing video information compared to single frame prediction. 
	The mean intersection over union (mIoU) metric on the Cityscapes validation set increases from \unit{45.2}{\%} for the single frames to \unit{57.9}{\%} for video data after implementing the ConvLSTM to propagate features trough time on the ESPNet. 
	Most importantly, inconsistency decreases from \unit{4.5}{\%} to \unit{1.3}{\%} which is a reduction by \unit{71.1}{\%}. 
	Our results indicate that the added temporal information produces a frame-to-frame consistent and more accurate image understanding compared to single frame processing.
	Code and videos are available at \small \url{https://github.com/mrebol/f2f-consistent-semantic-segmentation}
\end{abstract}
\vspace{-9pt}

\section{Introduction}  
We address the task of semantic segmentation which assigns a semantic class for each pixel in an image.
Our focus is on the computation of semantic segmentation for multiple consecutive images, referred to as frames, in a video sequence. 
Consecutive video frames contain similar information, because they capture a scene which only changes slightly. 
Therefore, the semantic segmentation of consecutive frames is similar as long as motion between frames does not increase significantly.
For example, consider a street scene recorded by a camera mounted on a vehicle in which we observe a street sign. 
If the frame rate is large enough, we will observe the street sign in multiple images as the vehicle passes by. 
In this example, the goal of this work would be to consistently detect the street sign as such in all frames in which the sign appears. 
Single frame algorithms often fail at achieving this task.  
In general, we aim for temporally consistent segmentation of all semantic classes throughout a video sequence. 

\begin{figure}[t]
	\vspace{-10pt}
	\centering
	\captionsetup[subfigure]{labelformat=empty, position=top, font={stretch=0}}
	\subfloat{\raisebox{-0.45in}{\rotatebox[origin=t]{90}{\small{Input}}}}\hspace*{1pt}%
	\subfloat[\small{Frame 1}]{\includegraphics[width=.48\columnwidth]{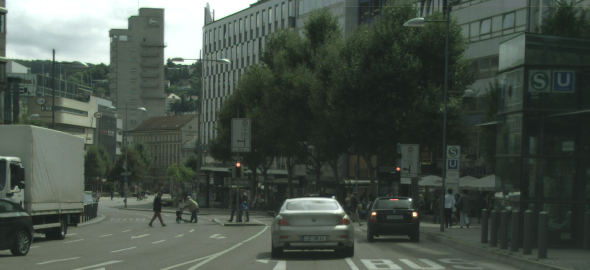} }
	\subfloat[\small{Frame 2}]{\includegraphics[width=.48\columnwidth]{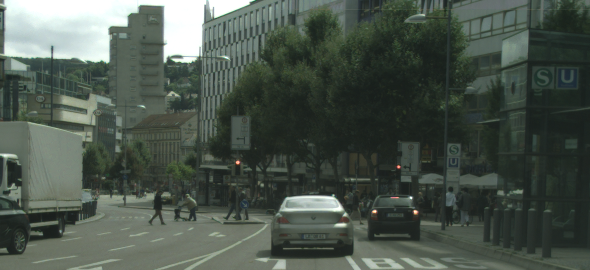}} 
	\vspace{-0.8em}
	\subfloat{\raisebox{-0.45in}{\rotatebox[origin=t]{90}{\small{ESPNet}}}}\hspace*{1pt}
	\subfloat{\includegraphics[width=.48\columnwidth]{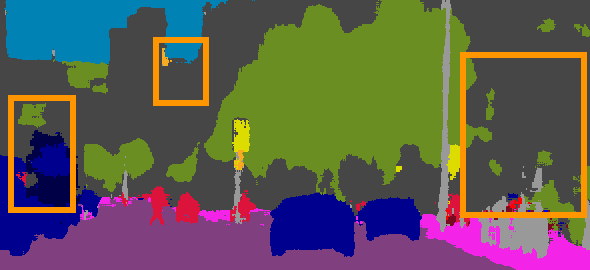}
		\includegraphics[width=.48\columnwidth]{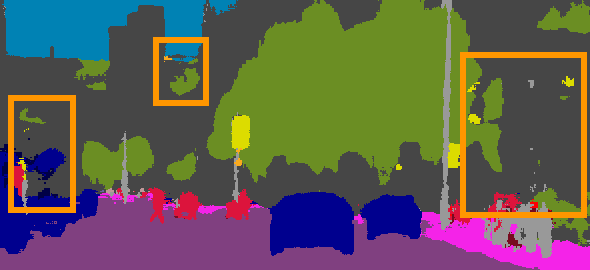}}
	\vspace{-0.8em}
	\subfloat{\raisebox{-0.45in}{\rotatebox[origin=t]{90}{\small{Our Model}}}}\hspace*{1pt}
	\subfloat{\includegraphics[width=.48\columnwidth]{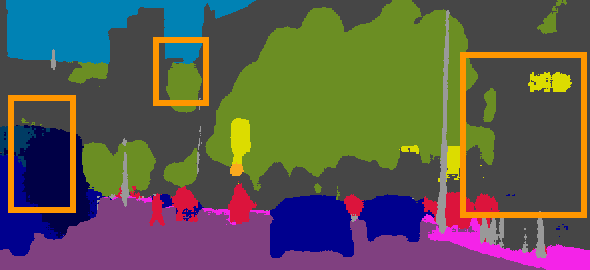}
		\includegraphics[width=.48\columnwidth]{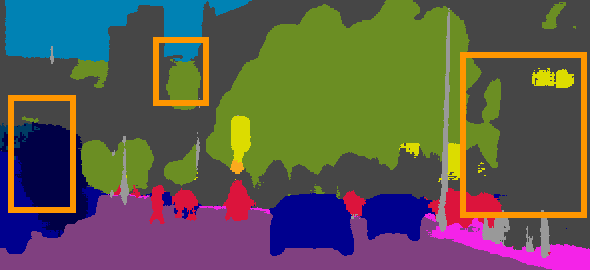}}
	\vspace{-5pt}
	\caption{Consistent Semantic Segmentation. The trained ESPNet \cite{espnet2018} model predicts temporally inconsistent semantic segmentation on two consecutive frames of the Cityscapes \cite{cityscapes2016} video data set (second row). 
		The semantic segmentation is color encoded and large inconsistencies are highlighted with orange boxes.
		The third row shows consistent results predicted by our model. 
		We reduce temporal inconsistencies by \unit{71}{\%}.}
	\label{fig:consistency}
	\vspace{-10pt}
\end{figure} 

Many state of the art computer vision algorithms process images individually \cite{imgclassnet2019, ssd15, DeeplabModels2018} and hence are not designed for video sequences. 
They do not consider the temporal dependencies which occur when segmenting a video semantically.  
If single frame convolutional neural networks (CNNs) predict semantic segmentation on video sequences, results can become temporally inconsistent because of illumination changes, occlusions and other variations. Figure \ref{fig:consistency} illustrates the differences between a temporally inconsistent prediction of video frames by a trained ESPNet \cite{espnet2018} and our consistent model. 

We address this issue by introducing methods which alter existing single frame CNN architectures such that their prediction accuracy benefits from having multiple frames of the same scene. 
Our method is designed such that it can be applied to any single frame CNN architecture. 
Potential applications include robotics and autonomous vehicles where video data can be recorded easily.
Since we aim for a real-life application scenario our method does not access future frames. 
Instead, we only utilize information from past frames to predict the current frame. 
We implement our online method on the lightweight CNN architecture ESPNet.
We include a recurrent neural network (RNN) layer into the ESPNet which allows past image features to be combined with current image features and thus computes consistent semantic segmentation over time. 
To train the parameters of our novel model for consistency, we introduce a inconsistency error term to our objective function. 
We verify our methods on a second architecture, which we name Semantic Segmentation Network (SSNet). 
The reason for the development of SSNet is to ensure that our methods do not only work on a specific CNN.  
We train the parameters of the two models on street scenes using supervised learning. The data is provided by the Cityscapes sequence \cite{cityscapes2016} and a synthetic data set which we generate from the Carla \cite{Carla2017} simulator. 
To avoid the large effort required to manually label video data, we use the pre-trained Xception model \cite{DeeplabModels2018} to predict highly accurate video semantic segmentation.

\section{Related Work} 
The best performances on semantic segmentation benchmark tasks such as PASCAL VOC \cite{Pascal15} and Cityscapes \cite{cityscapes2016} are reached by CNN architectures. 
Lightweight CNN architectures \cite{espnet2018, mobilenets2017, shuffleNet2017, Inceptionv4IA2016, resNext2016} have been developed to achieve high accuracy with low computational effort.
We select the highly efficient ESPNet \cite{espnet2018} as a basis for our work because it predicts semantic segmentation in real-time while maintaining high prediction accuracy. 
It uses point-wise convolutions together with a spatial pyramid of dilated convolutions \cite{dilConv2015}. 
The dilated convolutions allow the network to create a large receptive field while maintaining a shallow architecture.
Although ESPNet processes images fast and accurately, it lacks temporal consistency when predicting consecutive frames. 
Therefore, we extend the ESPNet and enforce video consistency. 
 
\paragraph{Video Consistency}
Kundu \etal \cite{Kundu2016} and Siddhartha \etal \cite{crf2018} base their work on the traditional graph cut \cite{Krahenbuhl2012, krahenbuhl2013} approach towards semantic segmentation. 
They extent the traditional 2D to a 3D CRF by adding a temporal dimension which allows them to predict temporally consistent semantic segmentation on video. 
Compared to our approach additional optical flow information needs to be computed and the size of the temporal dimension must be predefined in advance. 
This results in additional computation complexity and less flexibility when changing parameters such as the frame rate.  
Therefore, we decided to implement RNNs \cite{lstm97, convlstm2015, gru2014} which offer a more flexible approach towards processing video data. 

RNNs are trained to learn which features of past frames are relevant for current \cite{objLSTM2017, convgru2019} or future \cite{Srivastava2015, FutureSSNabavi2018} frames.
In general, it is not clear if LSTM, GRU or any other RNN architecture is superior \cite{GRUvsLSTM2014, LSTMvsGRU2015, LSTMsearch2017}. Depending on the application, one architecture might perform slightly better than the other \cite{GRUvsLSTM2014}. Variations through modifying the proposed architectures might work even better in some cases \cite{LSTMvsGRU2015}. The work of Jozefowicz \etal \cite{LSTMvsGRU2015} shows the importance of the elements inside an RNN cell.

Lu \etal \cite{objLSTM2017} use the plain LSTM to associate objects in a video. To enforce a frame-to-frame consistent prediction, they use an association loss during the training of the LSTM.
Similarly, we implement a ConvLSTM and an inconsistency loss to tackle semantic segmentation. 
We place the ConvLSTM on different image feature levels in our architecture as suggested by \cite{FutureSSNabavi2018, convgru2019}.

\section{Consistent Video Semantic Segmentation} 
In this section, we introduce our methods towards frame-to-frame consistent semantic segmentation. We present different architecture to propagate features through time. To train the architectures for temporal consistency, we extend the cross entropy loss function with a novel inconsistency error term. 

\subsection{Temporal Feature Propagation} \label{sec:temporal-feature-propagation}
\begin{figure}[t]
	\begin{center}
		\includegraphics[width=.43\textwidth]{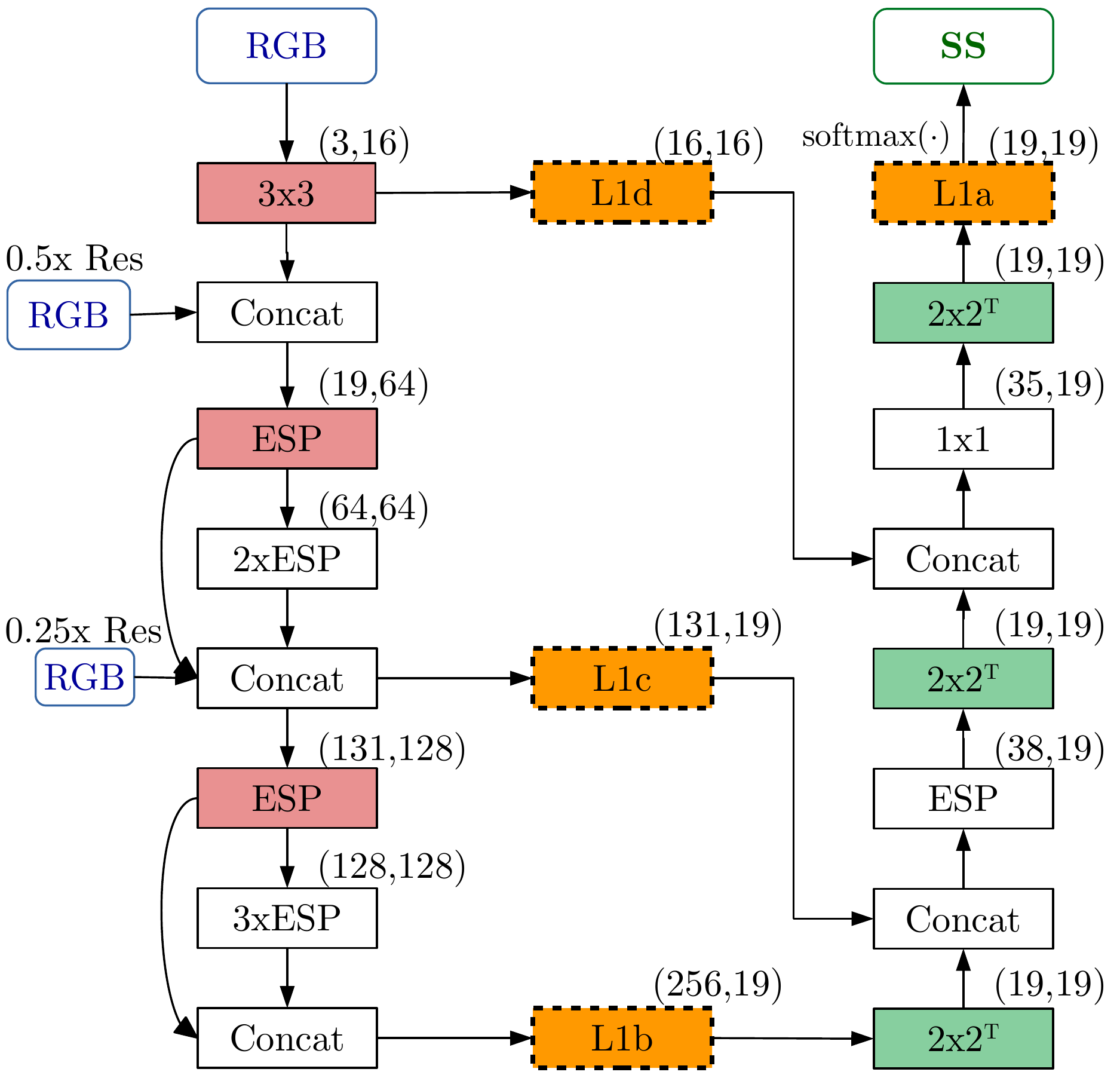}
	\end{center}
	\vspace{-0.9em}
	\caption{ESPNet with ConvLSTM. 
		Four different positions for including a ConvLSTM (orange) into the existing ESPNet architecture are depicted. Dashed boxes indicate that only one ConvLSTM is present in a single architecture. L1b, L1c and L1d replace $1 \times 1$ channel reduction convolutions while L1a adds an additional layer to the architecture of the original ESPNet. Red boxes indicate a spatial dimensionality reduction by the factor two, while green boxes indicate a spatial dimensionality increase of two.}
	\label{fig:ESPNet_L}
\end{figure}

The propagation of image features from the past to the current time step allows the neural network to make predictions based on time sequences. 
We prefer the ConvLSTM \cite{convlstm2015} cell for this dense prediction task. 
Compared to the fully connected LSTM, it removes unnecessary connections. 
For instance, the connection of features from the top left corner of the previous frame to features of the bottom right corner of the current frame is not needed. 
We assume that if we ensure consistency locally by the convolution operator, we will generate overall results which are consistent, as long as motion between frames can be detect in the local window.  
Therefore, we need to choose the filter size large enough to allow the ConvLSTM to detect local consistencies and motion between frames without explicit optical flow information. 
Furthermore, the ConvLSTM allows us to process images at different resolutions and reduces the number of parameters significantly compared to the fully connected LSTM. 
The definition of ConvLSTM cell is shown in \cite{convlstm2015}. 
We use two different networks in which we include the ConvLSTM cell. 
First, we introduce the Video SSNet (VSSNet) architecture which consists of six layers of $3 \times 3$ convolutions with dilation rates $\{1,1,2,2,4,4\}$ and 64 channels. 
Compared to the SSNet, we replace the last convolutional layer with a ConvLSTM in the VSSNet. 
Second, we also extend the ESPNet \cite{espnet2018} with a ConvLSTM layer. 
Although it would be reasonable to propagate features at every layer of a CNN architecture this is not feasible because of fast growing computational complexity. 
Figure \ref{fig:ESPNet_L} shows the ESPNet architecture with four possible positions for the ConvLSTM. 
The proposed architectures are enumerated alphabetically from ESPNet\_L1a to ESPNet\_L1d, starting with the ConvLSTM at the highest feature level which means that it is located closest to the output layer. 
Besides the ConvLSTM layer, we implement two ESP modules at the first spatial level and three ESP modules at the second spatial level, which is the simplest configuration 
introduced in \cite{espnet2018}. All other aspects of the ESPNet architecture remain unchanged.

\subsection{Temporal Consistency Loss}
Our second building block to enforce consistency is an additional error term in our loss function.  
The resulting loss function $\mathcal{L}(\cdot)$ is defined as
\begin{equation} \label{eqn:total_loss} \small
\mathcal{L}(\mat{S},\mat{P}) = \lambda_\text{ce} \mathcal{L}_\text{ce}(\mat{S},\mat{P}) + \lambda_\text{incons} \mathcal{L}_\text{incons}(\mat{S},\mat{P}) ,
\end{equation}
where $\mat{S}\in \mathbb{S}^{T \times M \times N}$ contains the semantic ground truth and $\mat{P}\in \mathbb{R}^{T \times M \times N \times |\mathbb{S}|}$ contains the predictions. 
The set $\mathbb{S}$ contains all semantic labels. 
We bound the dimensions by the sequence length $T$, the image dimensions $M \times N$ and the number of semantic labels $|\mathbb{S}|$.
The function $\mathcal{L}_\text{ce}(\cdot)$ computes the cross entropy loss and $\mathcal{L}_\text{incons}(\cdot)$ penalizes inconsistencies. The hyper-parameters  $\lambda_\text{ce}$ and $\lambda_\text{incons}$ are introduced to influence the balance between training with focus on prediction accuracy or consistency.

We define the inconsistency loss as 
{\small\begin{align} \label{eqn:incons3} 
&\mathcal{L}_\text{incons}(\mat{S},\mat{P}) = \frac{1}{\omega_\text{norm}(\mat{S})}  \! \! \sum_{t,m,n=1}^{T-1,M,N} \! \! \! \! \! \omega_\text{vcc}(\mat{S}, \mat{P}, t, m, n) \cdot \\
& \ \  \ \  \left(\sum_{s=1}^{|\mathbb{S}|}  \delta(\mat{S}_{t,m,n}=s) \cdot (\mat{P}_{t,m,n,s} - \mat{P}_{t+1,m,n,s})^2 \right) \notag , 
\end{align}}%
where $\delta(\cdot)$ refers to the indicator function defined as 
\begin{equation} \small 
\delta(\phi(\cdot)) = \begin{cases} 1 & \text{if } \phi(\cdot) \text{ is true } \\ 
0 & \text{else.} \end{cases} 
\end{equation}
The inconsistency loss penalizes pixels with different predictions in consecutive frames, which are already predicted correctly in at least one frame of the consecutive pair. 
This ensures that all other incorrect pixels are only affected by the cross-entropy loss. 
Additionally, $\delta(\mat{S}_{t,m,n}=s)$ selects only the correct semantic class for consistency enforcement. 
We normalize by the sum of pixels which are valid and consistent in the ground truth. 
This is achieved by 
\begin{equation} \label{eqn:cons-px}  \small
\omega_\text{norm}(\mat{S}) = \! \! \! \! \! \! \sum_{t,m,n=1}^{T-1,M,N} \! \! \! \!  \delta(\mat{S}_{t,m,n}\in\mathbb{S}) \cdot \delta(\mat{S}_{t,m,n} = \mat{S}_{t+1,m,n}), 
\end{equation} 
where the first factor checks for validity and the second one consistency in the ground truth. 
The boolean function $\omega_\text{vcc}(\cdot)$ ensures that only \emph{valid}, \emph{consistent} and \emph{correctly} (vcc) predicted pixels are affected by the following loss term.
{\small\begin{align} \label{eqn:vcr} 
\omega_\text{vcc}(\mat{S}, \mat{P}, t, m, n) \! = \, &\delta(\mat{S}_{t,m,n}\in\mathbb{S}) \ \cdot \\ 
& \delta(\mat{S}_{t,m,n} = \mat{S}_{t+1,m,n}) \ \cdot\notag\\ 
&\psi(\mat{S}_{t,m,n}, \mat{P}_{t,m,n}, \mat{S}_{t+1,m,n}, \mat{P}_{t+1,m,n}) ,\notag
\end{align}}%
where the first factor ensures validity, the second consistency and the third correct prediction in one of two consecutive images. The third factor is given by the boolean function $\psi(\cdot)$ which we define as
{\small\begin{align}
\psi(s_1, \vec{p}_1, s_2, \vec{p}_2) =	\min(&\delta(s_1 = \arg\max(\vec{p}_1)) \ + \notag \\
&\delta(s_2 = \arg\max (\vec{p}_2)), 1).
\end{align}}%
This function determines for a pixel at a certain position if at least one prediction in the consecutive image pair is correct. 
The input parameters are given by the two prediction vectors $\vec{p}_1, \vec{p}_2 \in \mathbb{R}^{|\mathbb{S}|} $ and the two ground truth labels $s_1, s_2 \in \mathbb{S}$ for any pixel position. All four parameters are retrieved from \mat{P} and \mat{S}. 

\begin{figure}[t]
	\centering
	\newcommand{\shiftleft}[2]{\makebox[0pt][r]{\makebox[#1][l]{#2}}}
	\captionsetup[subfigure]{labelformat=empty, position=top, font={small, stretch=0}}
	\subfloat{\raisebox{-0.45in}{\rotatebox[origin=t]{90}{\small{Frame 1}}}}\hspace*{3pt}%
	\subfloat[Prediction\vspace{-0.1em}]{\includegraphics[width=.48\columnwidth]{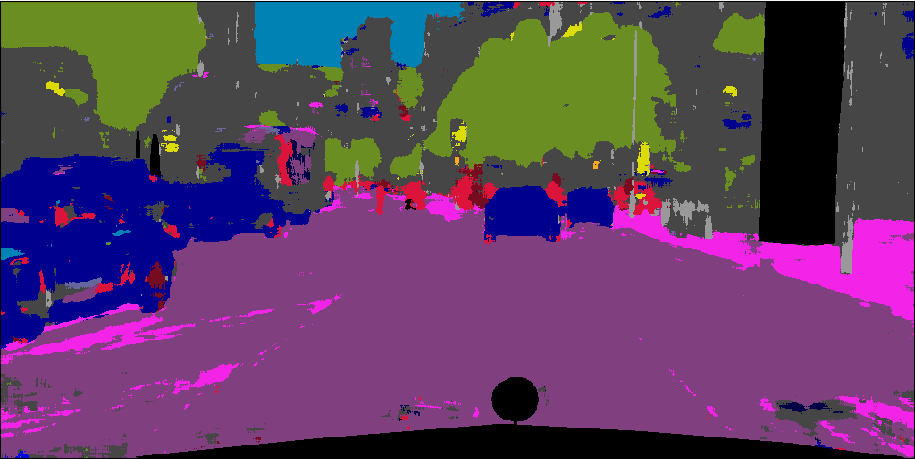} }
	\subfloat[Ground truth\vspace{-0.1em}]{\includegraphics[width=.48\columnwidth]{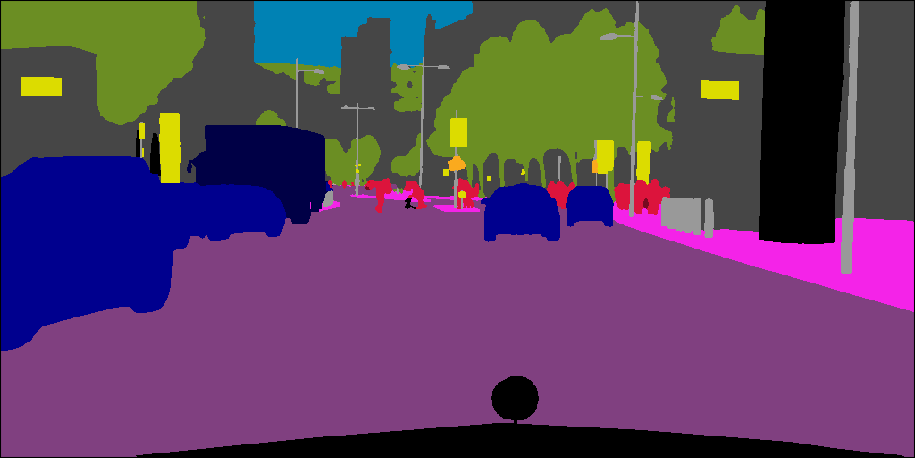} }%
	\vspace{-1.25em}
	\subfloat{\raisebox{-0.45in}{\rotatebox[origin=p]{90}{\small{Frame 2}}}}\hspace*{3pt}%
	\subfloat{\includegraphics[width=.48\columnwidth]{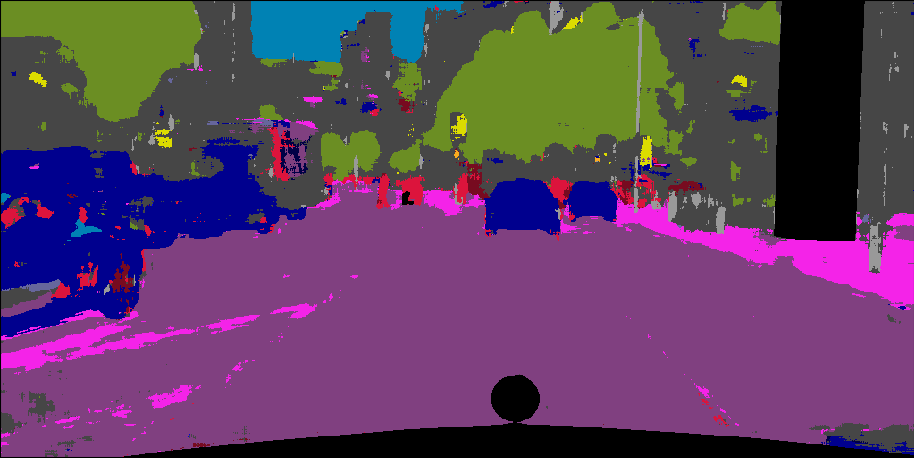} }
	\subfloat[]{\includegraphics[width=.48\columnwidth]{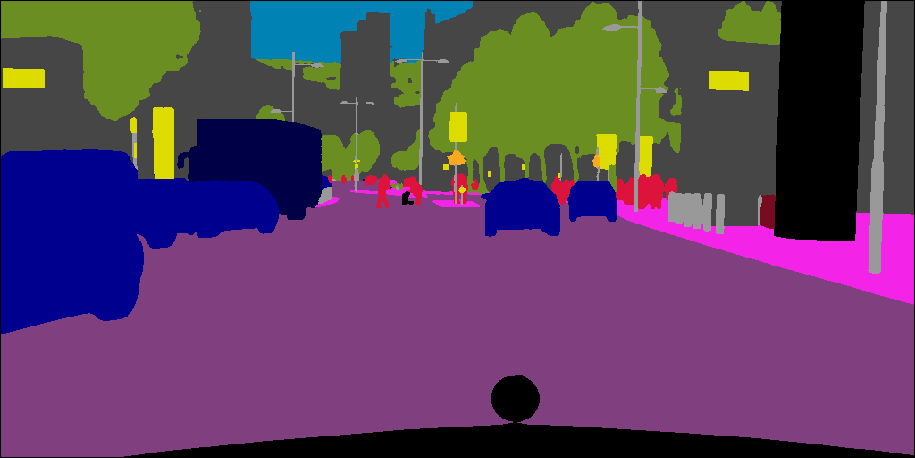} }%
	\vspace{-0.5em}
	\subfloat{\raisebox{-0.45in}{\rotatebox[origin=p]{90}{\small{\phantom{ABC 1}}}}}\hspace*{3pt}%
	\setcounter{subfigure}{0}%
	\captionsetup[subfigure]{labelformat=parens, position=top}%
	\subfloat[\footnotesize{Prediction inconsistency}\vspace{-0.3em}]{\includegraphics[width=.48\columnwidth]{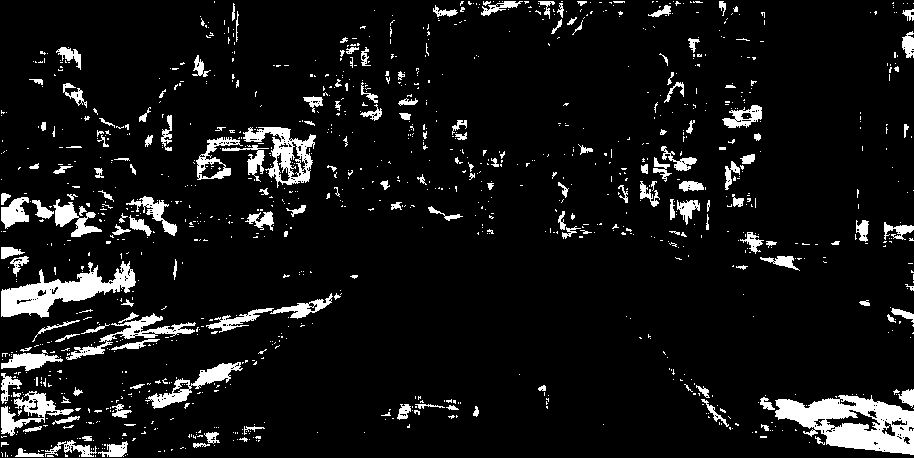}\label{fig:dilation:inconsistency} }%
	\subfloat[\footnotesize{Dilated GT scene change}\vspace{-0.3em}]{\includegraphics[width=.48\columnwidth]{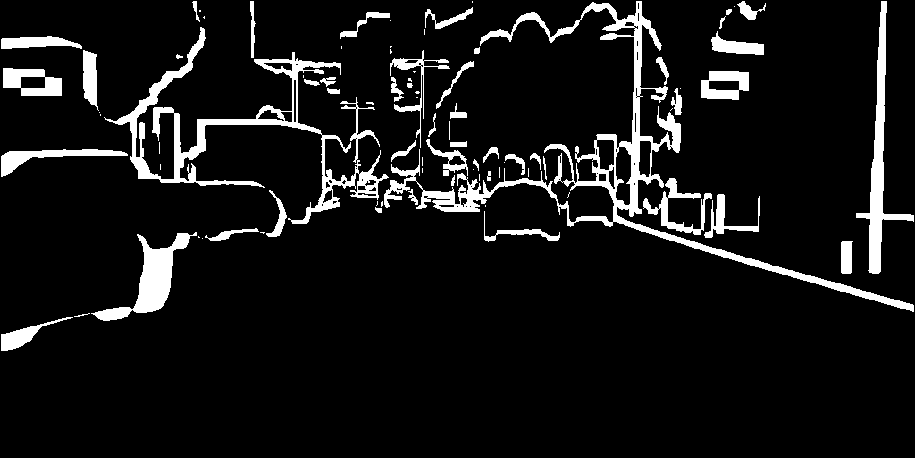}\label{fig:dilation:difference} }
	\vspace{-0.4em}
	\caption{Visualization of Inconsistencies. We compare prediction and ground truth at two different time steps. The white pixels in image \protect\subref{fig:dilation:inconsistency} are inconsistently predicted. 
	Image \protect\subref{fig:dilation:difference} shows pixels which change their label because of motion. Only black pixels in image \protect\subref{fig:dilation:difference} are affected by our inconsistency loss.}
	\label{fig:dilation}
\end{figure}

In Figure \ref{fig:dilation} we point out pixels which are affected by the inconsistency loss. In the bottom right of the prediction, the road (purple) is labeled inconsistently. For these pixels the function $\omega_\text{vcc}(\cdot)$ returns true and they are penalized by the inconsistency loss.  

\section{Experiments} 
First, we explain the generation of semantic video data with ground truth and show the impact of synthetic data. 
Second, we evaluate our proposed methods, \ie the feature propagation and the inconsistency loss. 

\paragraph{Architectures and data preparation} 
We use two models in our experiments, the ESPNet \cite{espnet2018} and the SSNet. 
We train the models on images with half and quarter Cityscapes resolution to reduce computational complexity. 
Comparisons between different configurations are always trained for the same number of epochs which is chosen high enough to allow for convergence of the configurations. 
We generate the pseudo ground truth for the sequence validation set with the Deeplab Xception model \cite{DeeplabModels2018}. 

\paragraph{Metrics and abbreviations}
The metrics which we use to compare our experiments are mean intersection over union (mIoU $\uparrow$), the percentage of correctly classified valid pixels (Acc $\uparrow$), the percentage of temporal consistently classified pixels (\Cons $\ \uparrow$) and the percentage of pixels which are temporally consistent but wrongly classified (\ConsW  $\ \downarrow$). The arrow pointing upwards $\uparrow$ indicates that a higher value is better, whereas the arrow pointing downwards $\downarrow$ indicates the opposite. Our \Cons\ and \ConsW\ metrics check all pixels which need to have the same label according to the ground truth, \ie black pixels in Figure \ref{fig:dilation:difference}.

\subsection{Data Generation}
An important part of our work is the generation of ground truth for a video data set.
We generate street scene video data with a pre-trained Deeplab Xception model \cite{DeeplabModels2018} and the Carla simulator \cite{Carla2017}.

\paragraph{Real world data} The semantic segmentation data sets of CamVid \cite{camvid2009}, Kitti \cite{kitti2012}, Cityscapes \cite{cityscapes2016} and Mapillary \cite{mapillary2017} do not provide ground truth for video data because of the large labeling effort required. 
Therefore, we use the Deeplab Xception model pretrained on the Cityscapes data set to generate pseudo ground truth labels for the Cityscapes sequence data set.
The reason why we prefer the Cityscapes dataset for video processing is that every 20th image of each sequence has annotated ground truth semantics. 
This allows for comparability with single frame results.

\paragraph{Synthetic data} 
Besides Cityscapes data, we also generate synthetic data with the Carla simulator \cite{Carla2017}.  
In total, we create 4680 scenes with 30 frames each. 
After training we evaluate on the Cityscapes sequence validation set. 
The quantitative results indicate that using about \unit{10}{\%} synthetic data slightly improves frame-to-frame consistency (Cons) from \unit{98.4}{\%} to \unit{98.5}{\%} for Cityscapes only training while mIoU remains at \unit{48.5}{\%}. 
When using more than \unit{20}{\%} of synthetic data, mIoU on the Cityscapes validation set declines significantly. 
We assume the reason for the decline is that only 9 of 19 semantic classes are covered by Carla data.
Nevertheless, we have shown that we can improve consistency by accurately labeled video semantic segmentation.  
For simplicity, we do not use the synthetic data set in other experiments.

\subsection{Feature Propagation Evaluation} \label{sec:feature_propagation}
First, we compare different ConvLSTM as well as inconsistency loss configurations. 
Finally, we combine the insights from the comparison to achieve the highest performance.

\begin{table}[]
	\setlength\extrarowheight{2pt} 
	\setlength{\tabcolsep}{2.6pt}
	\scriptsize
	\centering
	\begin{tabular}{l c|l|c c c c}
		\hline
		&\textbf{Category}&\textbf{Experiment}&\textbf{mIoU}&\textbf{Acc}&\textbf{\Cons}&\textbf{\ConsW}\\ \hline \hline
		& ESPNet &Single Frame &45.2&89.6&95.5&3.8\\ \hline \hline
		{\multirow{6}{*}{\rotatebox[origin=c]{90}{ConvLSTM}}}	
		&\multirow{2}{*}{\vspace{-3pt}\shortstack[1]{Convolution\\Type}} %
		& ESPNet\_L1a Std.& 46.5 & 89.4 & 97.6 & 5.4  \\ \cline{3-7}
		&& ESPNet\_L1a D.S.& 45.2& 89.0  & 97.2  & 5.5 \\ \cline{2-7} %
		&\multirow{4}{*}{\shortstack[1]{Position with\\Eq. Params.}} %
		& ESPNet\_L1a $7\!\times\!7$& 50.3 & 91.4 & 98.5 & 3.1  \\ \cline{3-7}
		&& ESPNet\_L1b $3\!\times\!3$& \textbf{52.0} & \textbf{91.5} & \textbf{98.7} & 3.2  \\ \cline{3-7}
		&& ESPNet\_L1c $5\!\times\!5$& 49.9 & 91.4 & 98.2 & 3.0  \\ \cline{3-7}
		&& ESPNet\_L1d $9\!\times\!9$& 50.1& \textbf{91.5}  & 98.3  & \textbf{2.9} \\ \hline 
	\end{tabular}
	\caption{ConvLSTM on ESPNet.
		Results on Cityscapes validation set. We compare the ESPNet trained with single frame images to different ConvLSTM configurations.%
	}\label{tab:lstm}
\end{table}	
\begin{table}[]
	\setlength\extrarowheight{2pt} 
	\setlength{\tabcolsep}{2.6pt}
	\scriptsize
	\centering
	\begin{tabular}{l c|l|c c c c}
		\hline
		&\textbf{Category}&\textbf{Experiment}&\textbf{mIoU}&\textbf{Acc}&\textbf{\Cons}&\textbf{\ConsW}\\ \hline \hline		
		{\multirow{5}{*}{\rotatebox[origin=c]{90}{Incons. Loss}}} %
		&\multirow{2}{*}{\vspace{-3pt}\shortstack[1]{Inconsistency\\Loss Func.}} %
		& Sq Diff True & 48.8 & 90.9 & 98.4 & {3.5}  \\ \cline{3-7}
		&& Abs Diff True & 48.6 & {90.9} & {98.6} & {3.5}  \\ \cline{3-7}
		\cline{2-7}
		&\multirow{3}{*}{\shortstack[1]{Inconsistency\\$\lambda$}} %
		& $\lambda_\text{incons} = 0$ & {49.0} & {90.9} & 98.0 & {3.4}  \\ \cline{3-7} %
		&& $\lambda_\text{incons} = 10$ & 48.8 &  {90.9} &98.4 & 3.5  \\ \cline{3-7}
		&& $\lambda_\text{incons} = 100$ & 46.3 & 90.4 & {98.6} & 3.7  \\ \hline \hline
		&\multirow{2}{*}{\vspace{-3pt}\shortstack[1]{\textbf{Comb. Results}\\\textbf{ESPNet\_L1b}}} & \textbf{On Val. Set}     & 57.9 & \textbf{93.0}   & \textbf{98.7} & \textbf{2.7}  \\ \cline{3-7} %
		&& \textbf{On Test Set}     & \textbf{60.9} & -   & - & -  \\ \hline %
	\end{tabular}
	\caption{Top: Inconsistency Loss.
		We vary parameters of the loss function. Note that the inconsistency loss results cannot be compared directly to Table \ref{tab:lstm} because we only train the LSTM parameters for faster convergence. 
		Bottom: Combined Results.
		The last two rows show the best results we are able to produce on Cityscapes validation and test set by combining the insights of our experiments. %
	}\label{tab:incons}
\end{table}

\paragraph{ConvLSTM on VSSNet} 
Training the VSSNet with ConvLSTM and inconsistency loss results in \unit{44.6}{\%} mIoU, \unit{89.9}{\%} Acc, \unit{97.7}{\%} \Cons \ and \unit{4.7}{\%} \ConsW. 
The results indicate that we are able to improve accuracy and consistency significantly, compared to the SSNet architecture trained with single frames which only achieves \unit{39.9}{\%} mIoU and \unit{94.4}{\%} \Cons.
After we have shown improvements on the VSSNet, we implement the following experiments on the ESPNet.  

\begin{figure*}
	\centering
	\setlength{\tabcolsep}{0pt}
	\renewcommand{\arraystretch}{0.5}
	\captionsetup[subfigure]{labelformat=empty, position=top}
	\newcommand{\shiftleft}[2]{\makebox[0pt][r]{\makebox[#1][l]{#2}}}
	\newcommand{\myvspace}{\vspace{-0.4em}}
	\subfloat{\shiftleft{10pt}{\raisebox{-.33in}{\rotatebox[origin=t]{90}{\small{Input}}}}}
	\subfloat[
		Frame 1
		\vspace{-.4em}]{\includegraphics[width=.2\textwidth]{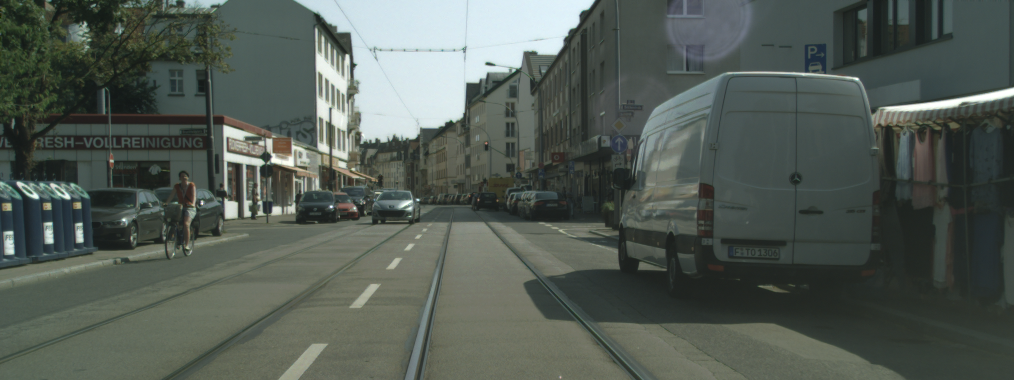} }
	\subfloat[
		Frame 2
		\vspace{-.4em}]{\includegraphics[width=.2\textwidth]{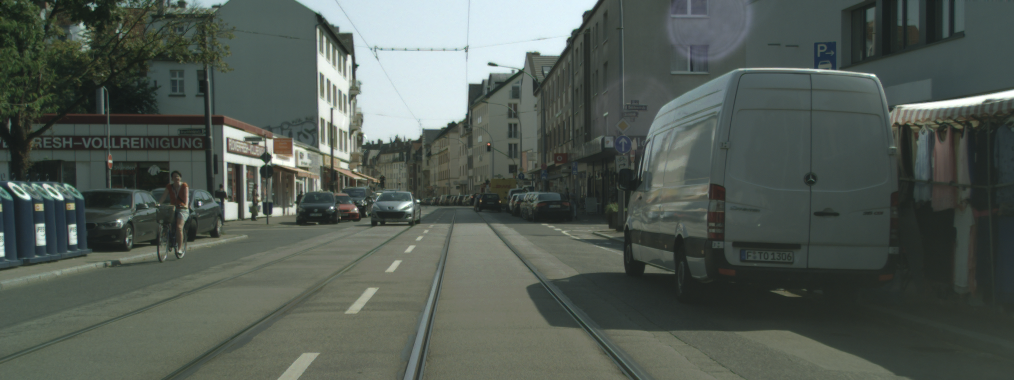} }
	\subfloat[
		Frame 3
		\vspace{-.4em}]{\includegraphics[width=.2\textwidth]{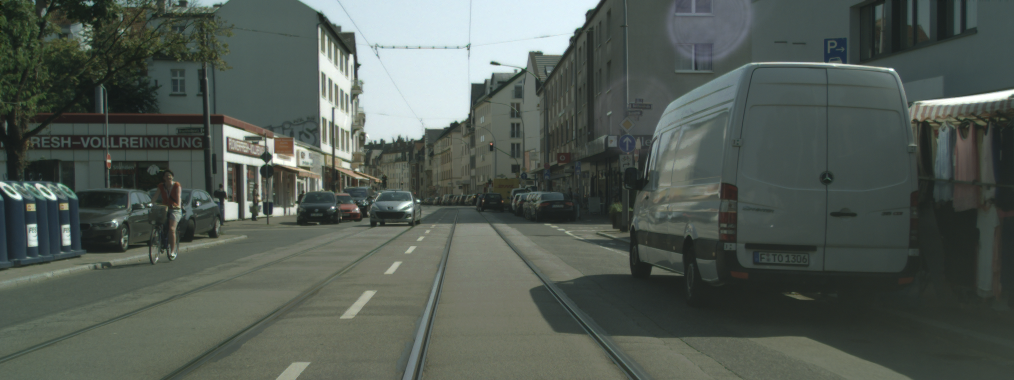} }
	\subfloat[
		Frame 4
		\vspace{-.4em}]{\includegraphics[width=.2\textwidth]{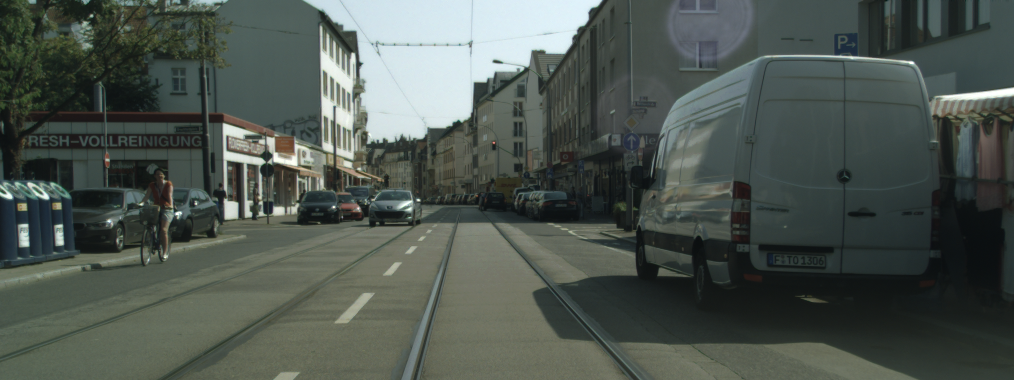} }
	\subfloat[
		Frame 5
		\vspace{-.4em}]{\includegraphics[width=.2\textwidth]{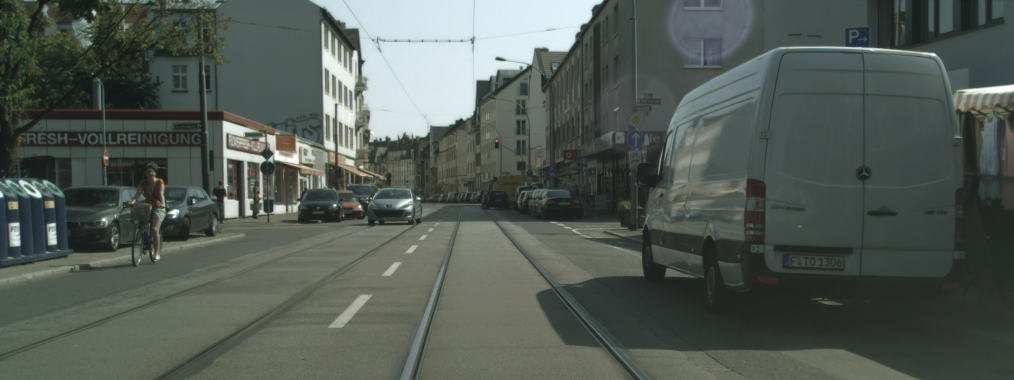}}
	
	\myvspace
	\subfloat{\shiftleft{10pt}{\raisebox{-.33in}{\rotatebox[origin=t]{90}{\small{GT}}}}}
	\subfloat{\includegraphics[width=.2\textwidth]{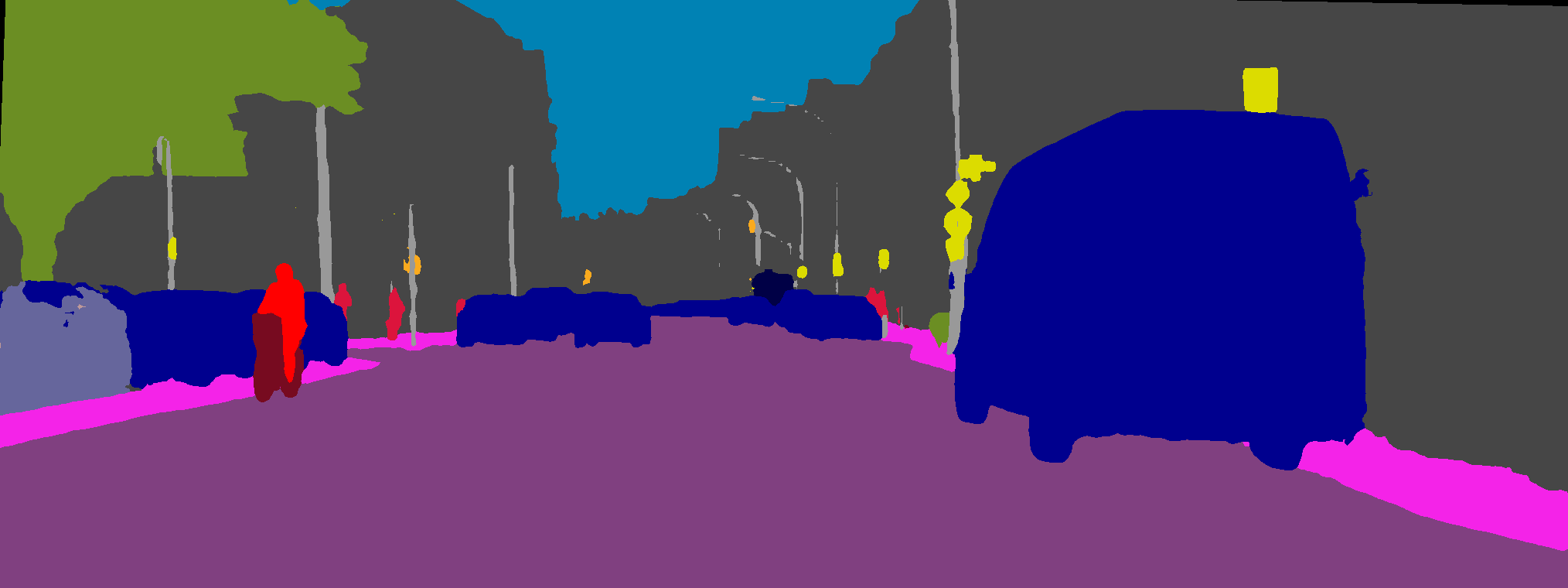}
		\includegraphics[width=.2\textwidth]{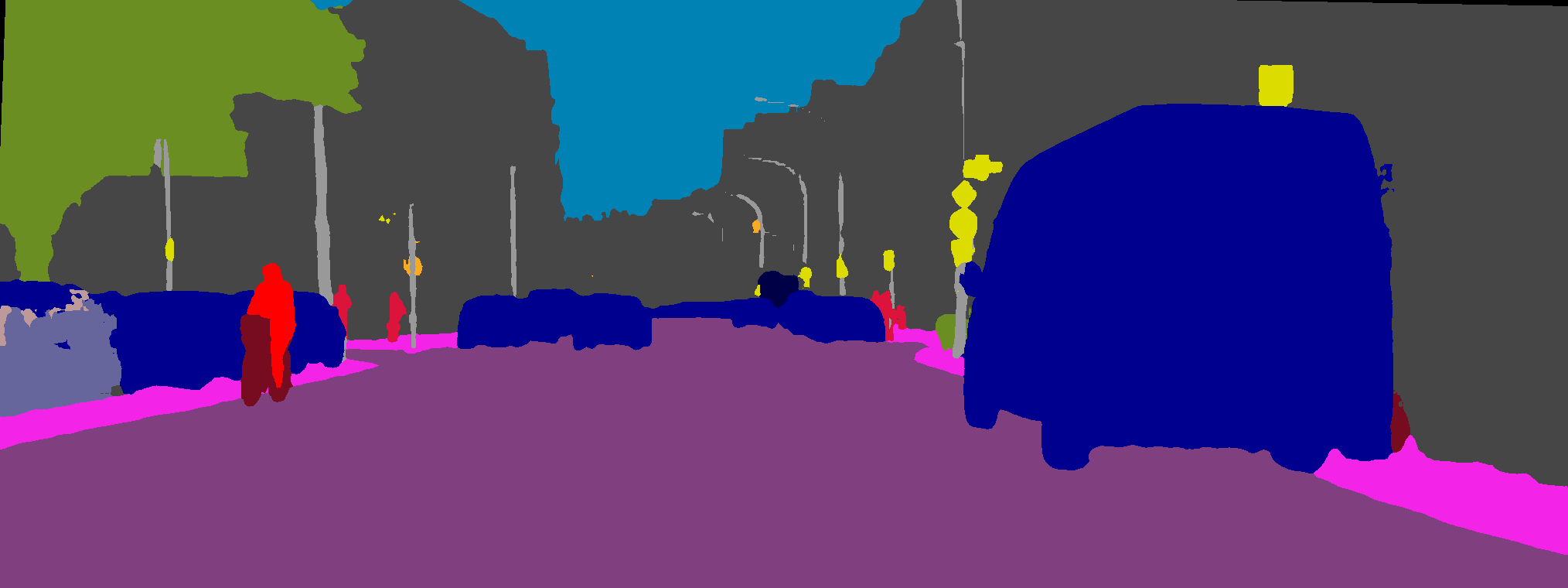}
		\includegraphics[width=.2\textwidth]{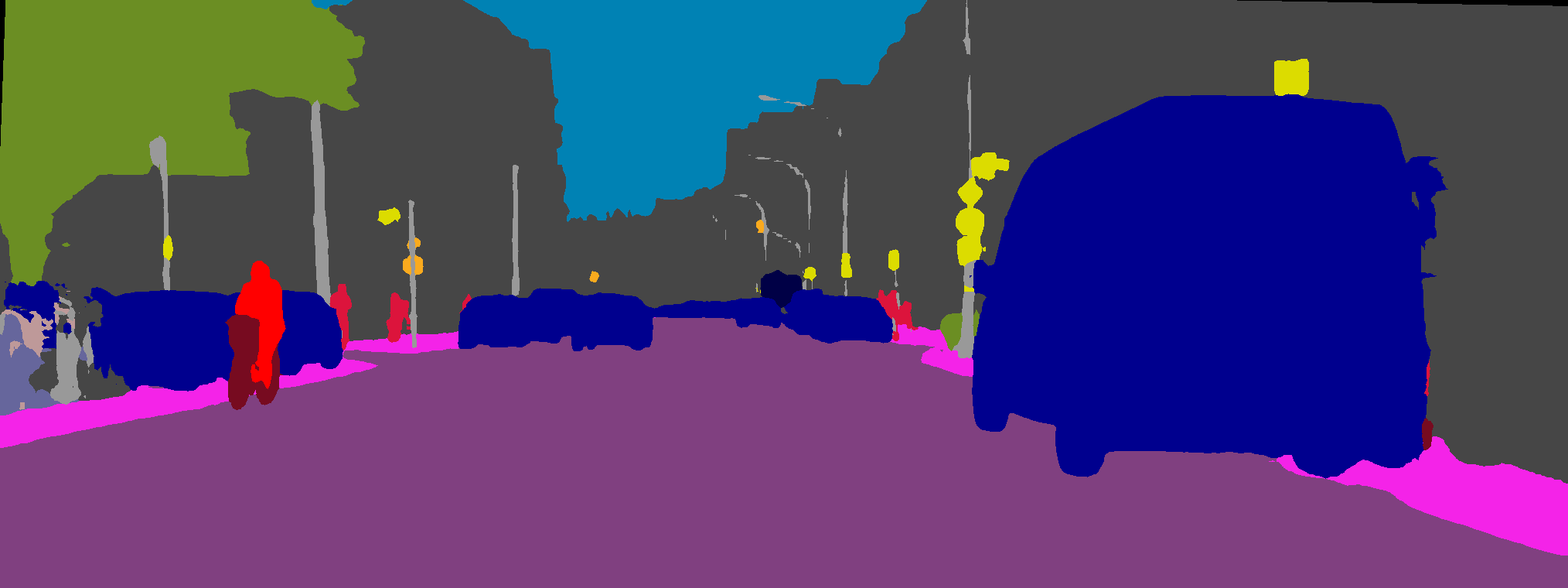}
		\includegraphics[width=.2\textwidth]{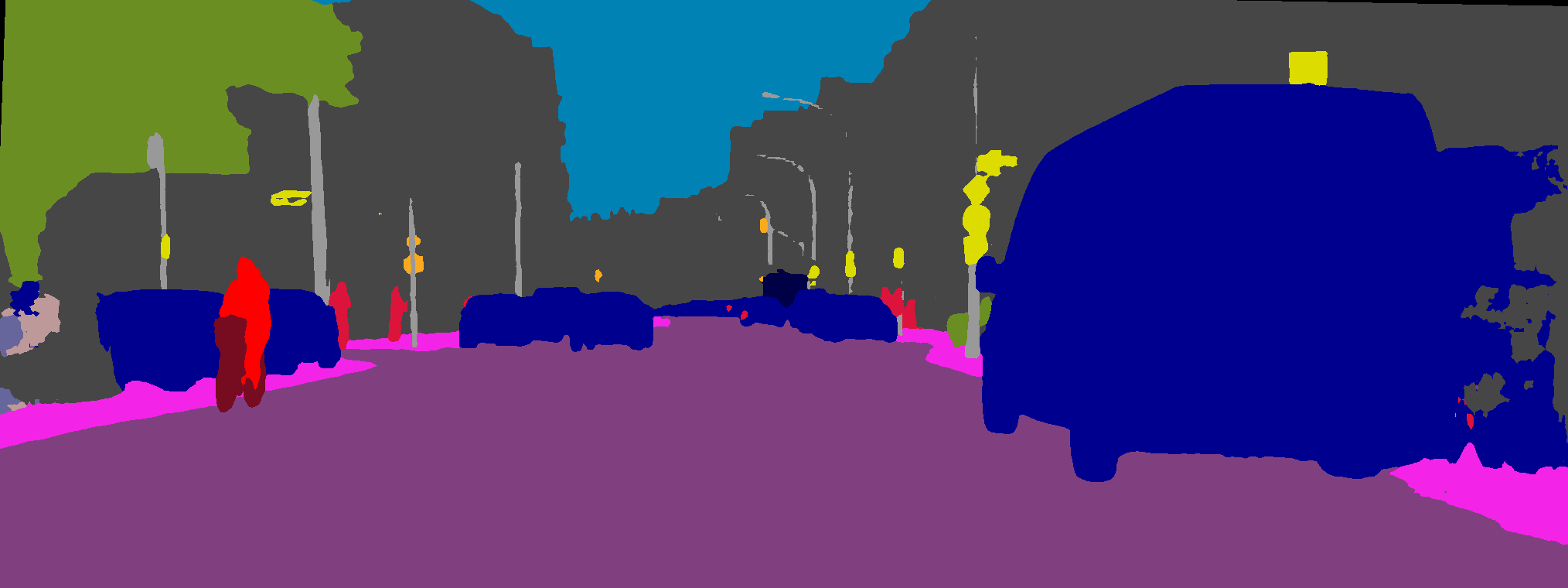}
		\includegraphics[width=.2\textwidth]{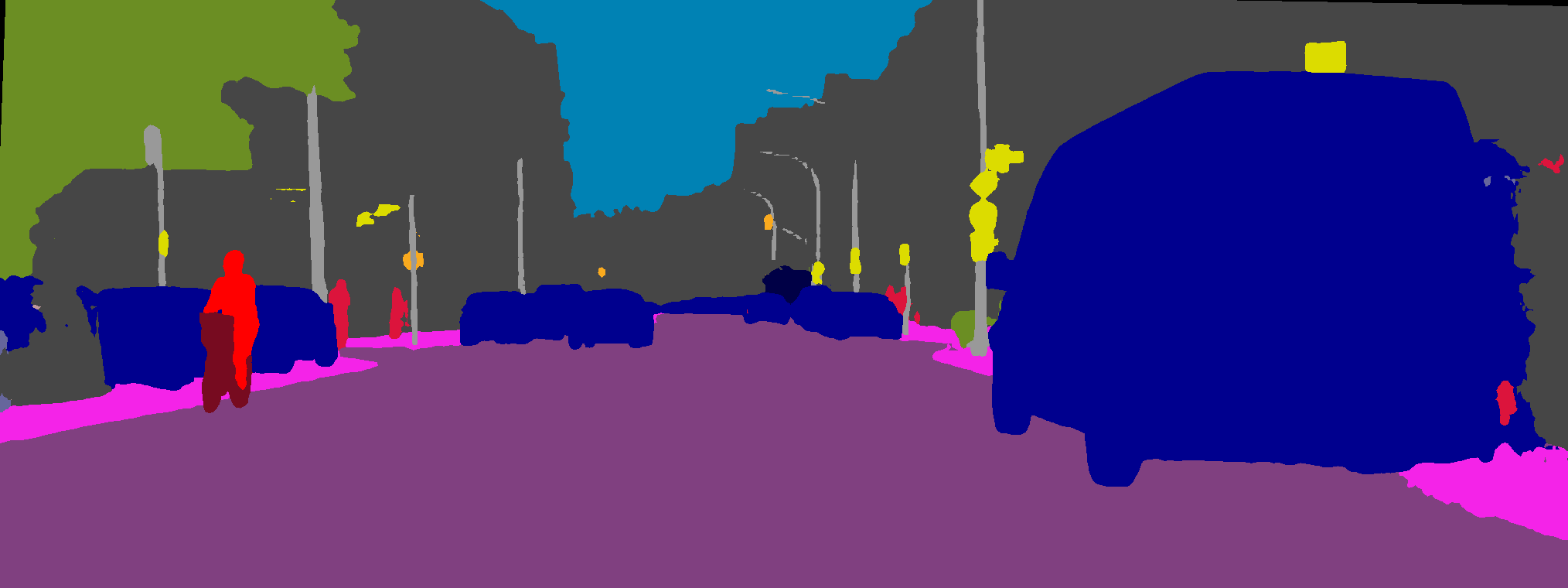}}
	
	\myvspace
	\subfloat{\shiftleft{10pt}{\raisebox{-.6in}{\rotatebox[origin=t]{90}{\small{ESPNet Sgl Train}}}}}
	\subfloat{\begin{tabular}[b]{c}\includegraphics[width=.2\textwidth]{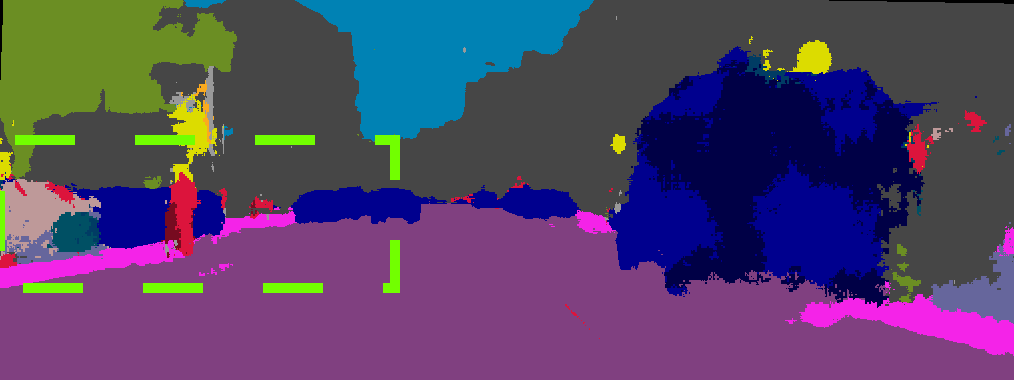}\vspace{-0.1em}\\
			\includegraphics[width=.2\textwidth]{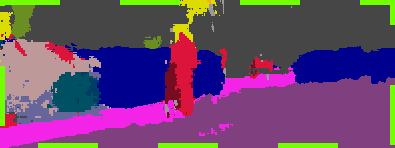}\end{tabular} }
	\subfloat{\begin{tabular}[b]{c}
			\includegraphics[width=.2\textwidth]{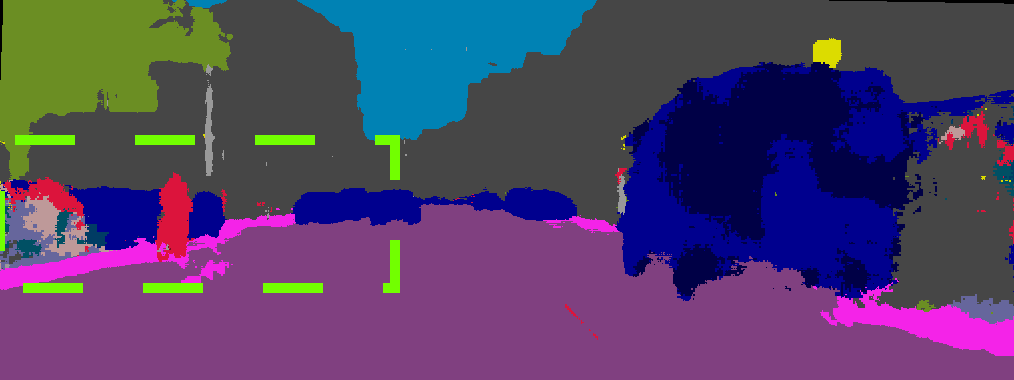}\vspace{-0.1em}\\
			\includegraphics[width=.2\textwidth]{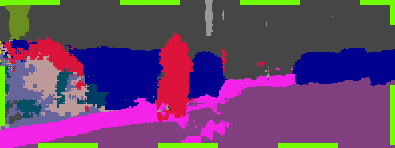}\end{tabular} }
	\subfloat{\begin{tabular}[b]{c}
			\includegraphics[width=.2\textwidth]{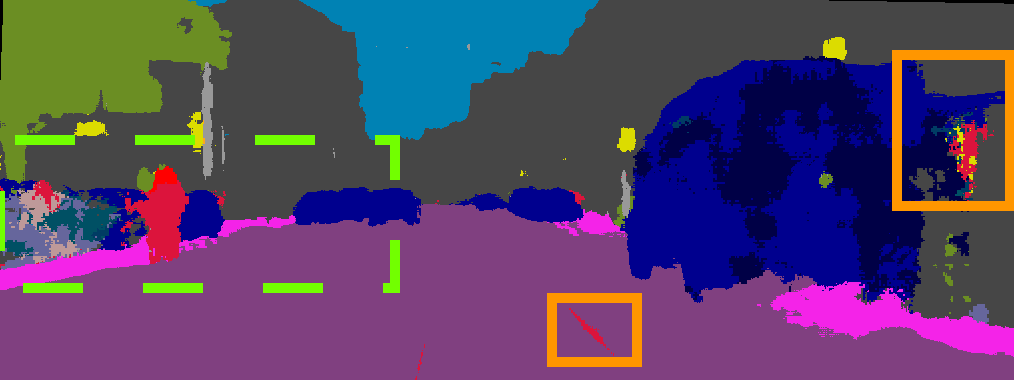}\vspace{-0.1em}\\\includegraphics[width=.2\textwidth]{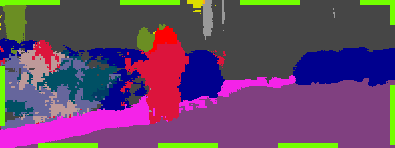}\end{tabular} }
	\subfloat{\begin{tabular}[b]{c}
			\includegraphics[width=.2\textwidth]{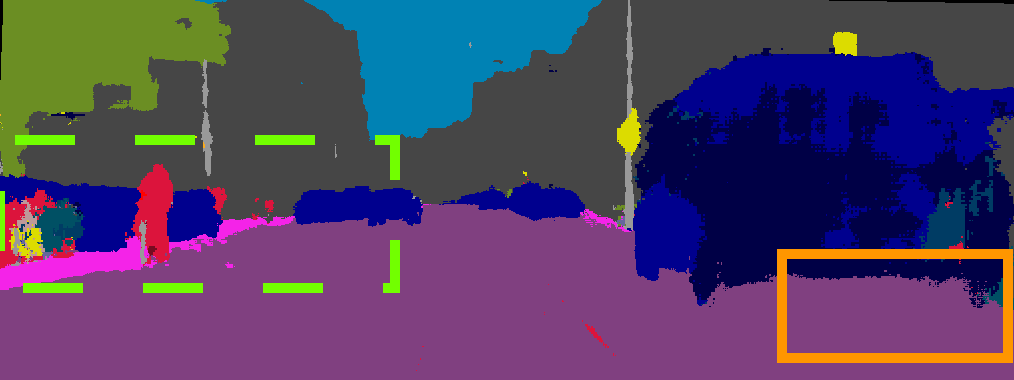}\vspace{-0.1em}\\\includegraphics[width=.2\textwidth]{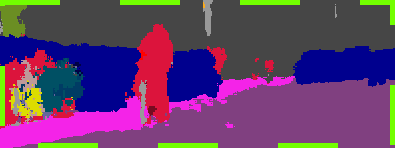}\end{tabular} }
	\subfloat{\begin{tabular}[b]{c}
			\includegraphics[width=.2\textwidth]{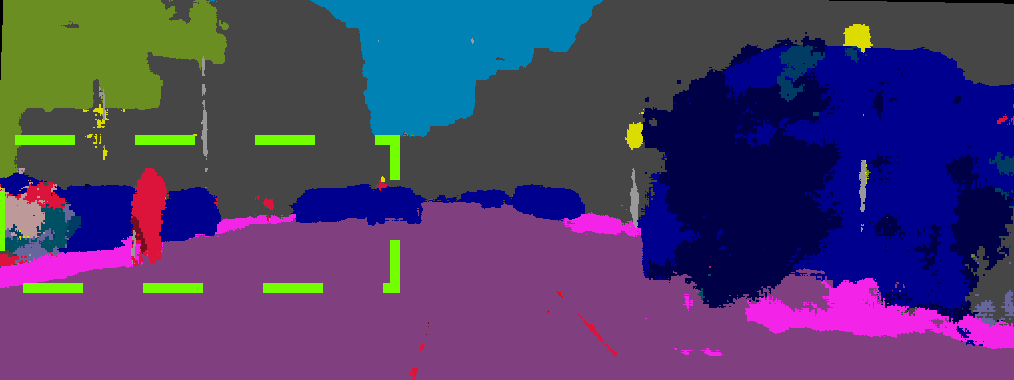}\vspace{-0.1em}\\\includegraphics[width=.2\textwidth]{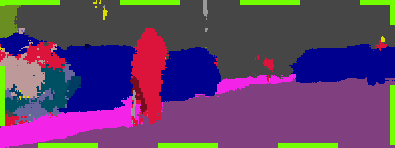}\end{tabular} }
	\vspace{-0.55em}
	\subfloat{\shiftleft{10pt}{\raisebox{-.6in}{\rotatebox[origin=t]{90}{\small{ESP.\_L1b}}}}}
	\subfloat{\begin{tabular}[b]{c}\includegraphics[width=.2\textwidth]{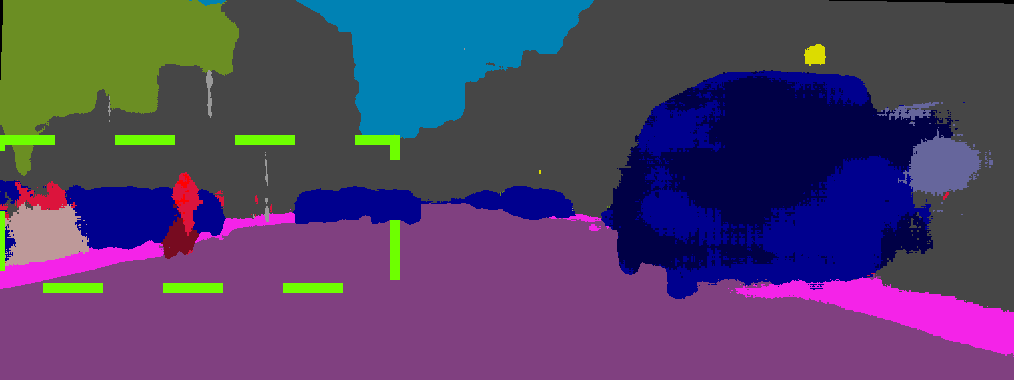}\vspace{-0.1em}\\
			\includegraphics[width=.2\textwidth]{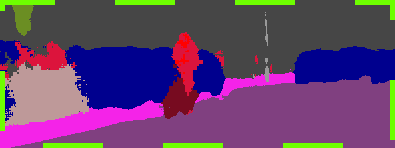}\end{tabular} }
	\subfloat{\begin{tabular}[b]{c}
			\includegraphics[width=.2\textwidth]{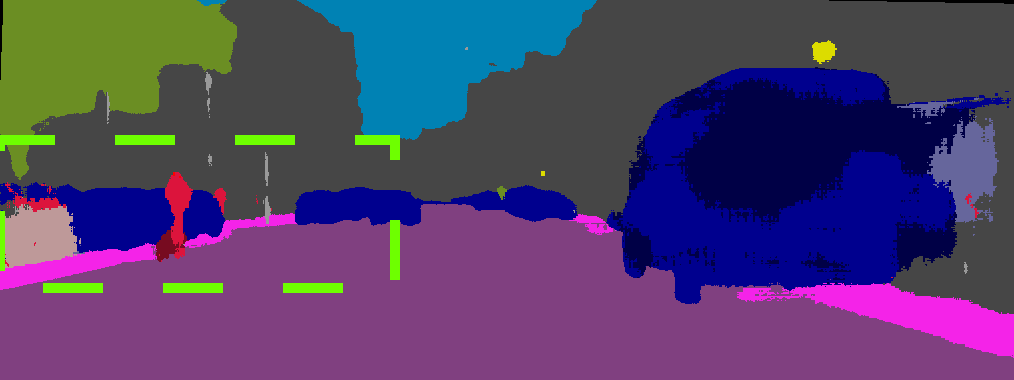}\vspace{-0.1em}\\
			\includegraphics[width=.2\textwidth]{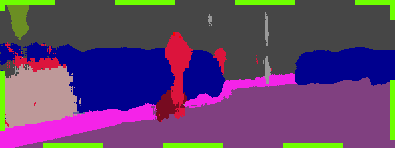}\end{tabular} }
	\subfloat{\begin{tabular}[b]{c}
			\includegraphics[width=.2\textwidth]{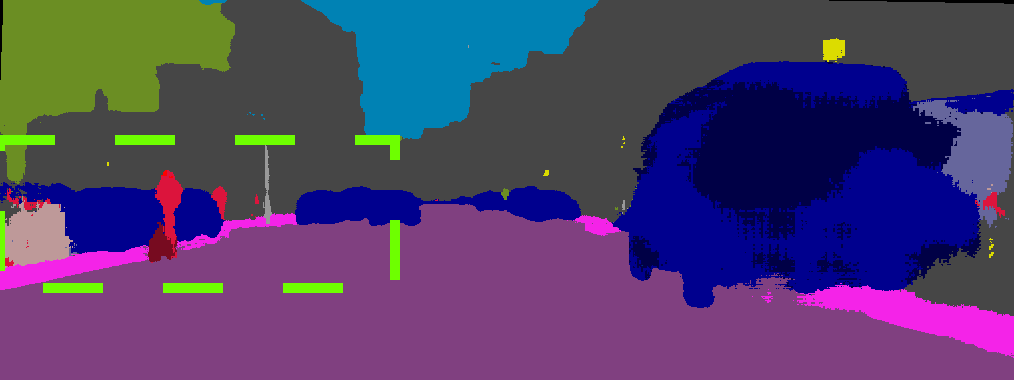}\vspace{-0.1em}\\\includegraphics[width=.2\textwidth]{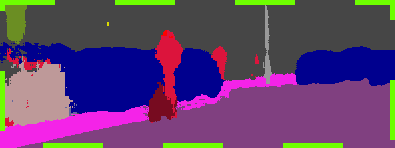}\end{tabular} }
	\subfloat{\begin{tabular}[b]{c}
			\includegraphics[width=.2\textwidth]{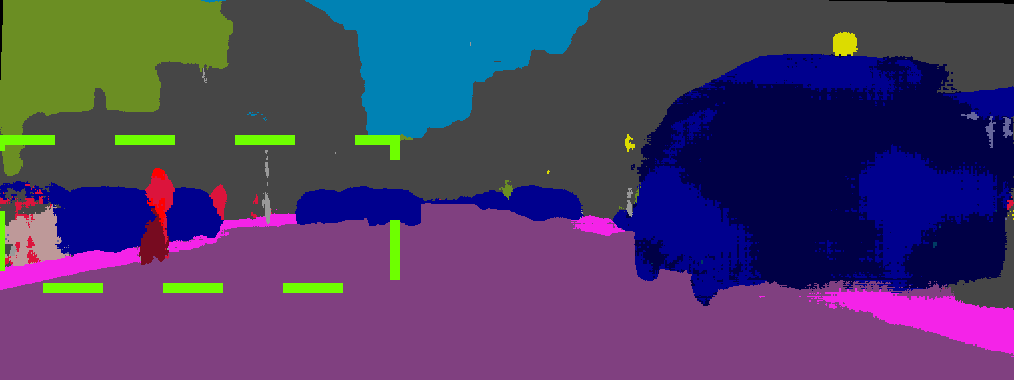}\vspace{-0.1em}\\\includegraphics[width=.2\textwidth]{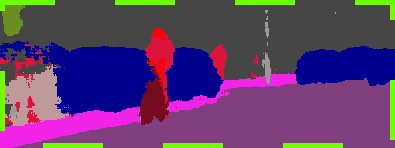}\end{tabular} }
	\subfloat{\begin{tabular}[b]{c}
			\includegraphics[width=.2\textwidth]{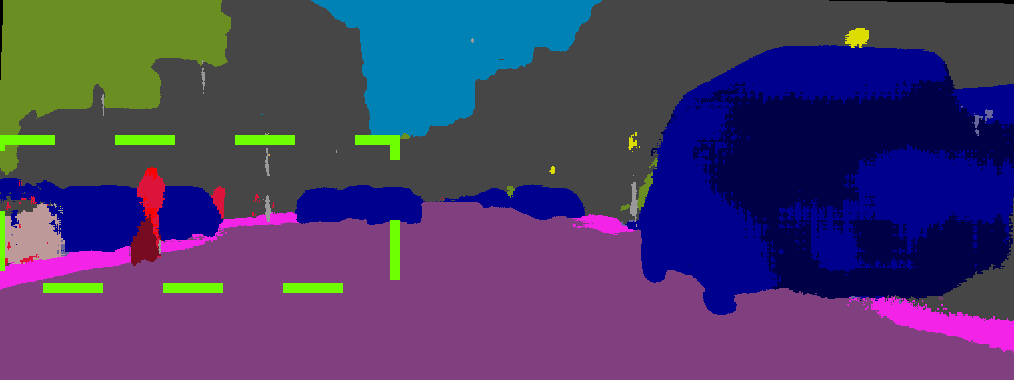}\vspace{-0.1em}\\\includegraphics[width=.2\textwidth]{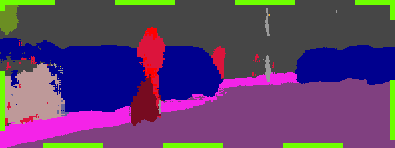}\end{tabular} }
	\vspace{-0.5em}
	\caption{Qualitative Results. A comparison between input data, DeepLab Xception ground-truth, single frame training and LSTM training on the ESPNet (top to bottom). The horizontal axis represents the time steps. Areas with inconsistent predictions are shown in detail and highlighted with green dashed boxes. Other inconsistencies are highlighted with orange boxes. The ESPNet with single frame training (Sgl Train) produces inconsistencies in the right, left and on the road segmentation. The ESPNet\_L1b predicts significantly more accurate and consistent results. }
	\label{fig:quality}
\end{figure*}  %

\paragraph{ConvLSTM configurations} 
We test different convolution types and positions of the ConvLSTM as proposed in Figure \ref{fig:ESPNet_L}. 
Table \ref{tab:lstm} shows the quantitative results of this comparison in the categories Convolution Types and Position with Equal Parameters. 
We compare the standard convolution operation with the depth-wise separable convolution inside the ConvLSTM on the ESPNet\_L1a architecture. 
Results show that the standard convolution inside the ConvLSTM produces better results for all four metrics. 

Furthermore, we evaluate the position of the ConvLSTM layer.
We choose the filter size such that all experiments have a similar number of parameters for a fair comparison.
This also ensures that the size of the receptive field at the layer is large enough to detect motion.  
The ESPNet\_L1b architecture clearly outperforms all other architectures in both consistency and accuracy. 
This suggests that it is more efficient to propagate high level image features. 
Additionally, we found that the Parametric ReLU ($\prelu$) performs better than the $\tanh$ activation function inside the ConvLSTM.
Therefore, results are reported implementing the $\prelu$ activation function.

\paragraph{Inconsistency loss}
We test different inconsistency loss configurations on the ESPNet\_L1b architecture because this model delivered the best results in previous experiments. 
Table \ref{tab:incons} contains the quantitative results. 
We only train the LSTM parameters to allow for fast comparison of multiple models. The other parameters of the model are pretrained, but do not receive updates after the LSTM cell is added. Consequently, the scores are slightly lower than in Table \ref{tab:lstm}. 
Substituting the squared difference loss inside Equation \eqref{eqn:incons3} with the absolute difference produces similar results. 
We observe that the hyper-parameter $\lambda_\text{incons} = 10$ provides a good trade-off between accuracy and consistency when using the squared difference loss function. The increase in consistency by 0.4 percentage points is noticeable when comparing the qualitative results. 
We set the other hyper-parameter $\lambda_\text{ce} = 1$ for all of our experiments.

\paragraph{Combining the findings}
In order to achieve the best results with ESPNet\_L1b, we train the model in multiple phases. 
We use the squared difference inconsistency loss on correctly predicted classes with $\lambda_\text{incons} = 10$ and a $5 \times 5$ convolution inside the ConvLSTM. 
The quantitative results are shown at the bottom of Table \ref{tab:incons}.
When training with the weighted cross entropy loss and data augmentations as proposed in \cite{espnet2018} the official Cityscapes server reports \unit{60.9}{\%} mIoU on the single frame test set.
Our method reaches slightly higher accuracy and significantly better temporal consistency while using a similar number of parameters as Metha \etal \cite{espnet2018}.

\section{Conclusion}
We have shown that we can improve temporal consistency and accuracy of semantic segmentation for two different single frame architectures by adding feature propagation and a novel inconsistency loss. 
On the ESPNet, consistency and mIoU improve from 95.5 to \unit{98.7}{\%} and from 45.2 to \unit{57.9}{\%}, respectively. This is equal to a reduction of inconsistencies by \unit{71.1}{\%} which can be observed immediately when watching a video sequence. 

Moreover, we found that it is best to forward features at a high level with a standard convolution within the ConvLSTM cell. The hyper-parameter in our novel inconsistency loss function can be used to prioritize between consistency and accuracy. We also improve consistency slightly by adding synthetic data generated by the Carla simulator.  

In future experiments we are interested in comparing other methods of adding the information from past frames to the current prediction. 
We also need to generate synthetic data such that it contains semantics of all validation classes to increase overall consistency and accuracy.

{\small
\bibliographystyle{ieee}
\bibliography{acvrw}
}

\input{f2fcss_suppl.tex}

\end{document}

%% file: f2fcss_suppl.tex
\clearpage
\newpage

\twocolumn[
\begin{center}
	{\Large
		\vskip0.5cm
		\textbf{Frame-To-Frame Consistent Semantic Segmentation\\ --- Supplementary Material ---}
		\vskip1cm}
\end{center}
]

\appendix

\section{Data Generation}
We give a detailed description of the data generation with the Carla simulator and evaluate the quality of the pseudo ground truth created by the DeepLab model. 

\subsection{Data Generation with the Carla Simulator}
We start the Carla server and control the simulation on the client via the Python API. 
We capture synthetic street scenes on the five maps Town1 to Town5. 
Town1 to Town4 are used to generate training data and Town5 is used to generate test data. 
Table \ref{tab:carla-gen} summarizes the generation configuration.
After a map is loaded, we create traffic by simultaneously spawning vehicles on different locations of the map. 
The number of vehicles on the map is set according to the map size. 
It is shown for each map in the row \emph{Vehicles on Map} of Table \ref{tab:carla-gen}. 
Once traffic is created, the camera vehicle \emph{Master} is spawned to capture the scene. 
The recording starts between 3 to 10 seconds after \emph{Master} has spawned to avoid capturing the exact same scene multiple times. 
This time period also allows \emph{Master} to accelerate to the speed limit.
Every scene is recorded in asynchronous mode which allows to maintain the exact frame rate of 17 FPS. 
\emph{Master} records the scenes for the nine most natural appearing weather conditions which are shown in the row \emph{Wheater IDs} of Table \ref{tab:carla-gen}.

\subsection{Evaluation of DeepLab Pseudo Ground Truth}
In order to verify that the pseudo ground truth generated by the DeepLab model is suitable for temporal consistency training, we evaluate the model on 60 manually labeled images from Cityscapes video data set. 
This data has not been used during training of the model. 
It consists of six scenes with ten consecutive images for each scene. 
The DeepLab model reaches \unit[99.7]{\%} consistency on the manually labeled ground truth. Due to the high temporal consistency, we conclude that the DeepLab model is suitable for consistency training of other models.

\begin{table}
	\setlength\extrarowheight{2pt} 
	\footnotesize
	\centering
	\begin{tabular}{l||l|l|l|l||l}
		\hline %
		\textbf{Map Names}            & Tn1 & Tn2 & Tn3 & Tn4 & Tn5 \\ \hline
		\textbf{Master Spawn Pts}     & 130 & 70  & 100 & 100 & 120 \\ \hline
		\textbf{Vehicles on Map}      & 150 & 80  & 300 & 300 & 300 \\ \hline
		\textbf{Weather IDs}          & \multicolumn{5}{l}{1, 2, 3, 4, 7, 8, 9, 10, 11} \\ \hline
		\textbf{Time After Spawn}     & \multicolumn{5}{l}{{[}3, 10{]} sec}             \\ \hline%
	\end{tabular}
	\caption{Carla Generation Settings. The first four towns (Tn1 to Tn4) are used for training whereas Town 5 (Tn5) is used for validation. The number of Master Spawn Points (Pts) multipled by the number of Weather IDs, \ie 9, equals the number of scenes captured for a given map.}
	\label{tab:carla-gen}
\end{table}

\section{ConvLSTM Convolution Types}
In our paper we have shown that a ConvLSTM layer inside a CNN architecture leads to a highly consistent prediction over time.
In this section, we focus on different convolution types inside the ConvLSTM cell in more detail.
We also compare the number of learnable parameters for each convolution type.
We refer to the number of input and output channels for the ConvLSTM by the variables $C$ and $D$, respectively. 
The filter size of the ConvLSTM is represented by $P$ and $Q$.

\paragraph{Standard convolution}
The standard convolution inside the ConvLSTM takes the information of all input channels to compute each output channel. Therefore, we use it if we cannot semantically separate the input channels. It results in  
\begin{equation}
4\cdot (C + D) \cdot P\cdot Q \cdot D + 7\cdot D
\end{equation}
parameters.

\paragraph{Depthwise separable convolution}  
We use the depthwise separate ConvLSTM if the input channels can be semantically distinguished. This means that we only the information of a specific input channel to compute the output channel. In this case we keep the number of input channels equal to the number of output channels $D = C$.
This configuration needs 
\begin{equation}
8\cdot C\cdot P\cdot Q+7\cdot C
\end{equation}
parameters. Consequently, $8\cdot C\cdot (C-1)\cdot P\cdot Q$ parameters are saved compared to the standard convolution.

\paragraph{Depthwise separable convolution with weight sharing}  
If we have the same requirements as for the depthwise separable convolution and treat each input channels equally, we use the same weights for each channel. 
This reduces the number of parameters further to only 
\begin{equation}
8\cdot P \cdot Q + 7 .
\end{equation}

\begin{figure*}[h]
	\centering  
	\setlength{\tabcolsep}{0pt}
	\renewcommand{\arraystretch}{0.5}
	\captionsetup[subfigure]{labelformat=empty, position=top}
	\newcommand{\shiftleft}[2]{\makebox[0pt][r]{\makebox[#1][l]{#2}}}
	\newcounter{start}
	\setcounter{start}{22786}
	\newcounter{d}
	\setcounter{d}{1}
	\newcommand{\cropimg}[1]{\includegraphics[trim={330pt 200pt 200pt 10pt }, clip=true, width=.33\textwidth]{fig/#1}}  
	\subfloat{\shiftleft{10pt}{\raisebox{-.7in}{\rotatebox[origin=t]{90}{\small{Input}}}}}
	\subfloat[Frame 1]{
		\cropimg{quality/rgb/{\the\numexpr\value{start}+\value{d}*0}}}
	\subfloat[Frame 2]{
		\cropimg{quality/rgb/{\the\numexpr\value{start}+\value{d}*1}}}
	\subfloat[Frame 3]{
		\cropimg{quality/rgb/{\the\numexpr\value{start}+\value{d}*2}}}
	\vspace{-0.5em}
	\subfloat{\shiftleft{10pt}{\raisebox{-.77in}{\rotatebox[origin=t]{90}{\small{Ground Truth}}}}}
	\subfloat{
		\cropimg{quality/gt/{\the\numexpr\value{start}+\value{d}*0}}
		\cropimg{quality/gt/{\the\numexpr\value{start}+\value{d}*1}}
		\cropimg{quality/gt/{\the\numexpr\value{start}+\value{d}*2}}
	}
	\vspace{-0.5em}
	\subfloat{\shiftleft{10pt}{\raisebox{-.8in}{\rotatebox[origin=t]{90}{\small{ESPNet }}}}}
	\subfloat{
		\begin{tikzpicture}
		\node[anchor=south west,inner sep=0] at (0,0)		{\cropimg{quality/sf/{\the\numexpr\value{start}+\value{d}*0}}};
		\draw[color=highlight, line width=1.4pt,rounded corners=false] (114pt,82pt) rectangle (150pt,102pt);
		\draw[color=highlight, line width=1.4pt,rounded corners=false] (57pt,23pt) rectangle (85pt,50pt);
		\end{tikzpicture}
		\begin{tikzpicture}
		\node[anchor=south west,inner sep=0] at (0,0)		{\cropimg{quality/sf/{\the\numexpr\value{start}+\value{d}*1}}};
		\draw[color=highlight, line width=1.4pt,rounded corners=false] (114pt,82pt) rectangle (150pt,102pt);
		\draw[color=highlight, line width=1.4pt,rounded corners=false] (57pt,23pt) rectangle (85pt,50pt);
		\end{tikzpicture}
		\begin{tikzpicture}
		\node[anchor=south west,inner sep=0] at (0,0)		{\cropimg{quality/sf/{\the\numexpr\value{start}+\value{d}*2}}};
		\draw[color=highlight, line width=1.4pt,rounded corners=false] (114pt,82pt) rectangle (150pt,102pt);
		\draw[color=highlight, line width=1.4pt,rounded corners=false] (57pt,23pt) rectangle (85pt,50pt);
		\end{tikzpicture}
	}
	\vspace{-0.5em}
	\subfloat{\shiftleft{10pt}{\raisebox{-0.8in}{\rotatebox[origin=t]{90}{\small{ESPNet\_L1b}}}}}
	\subfloat{
		\begin{tikzpicture}
		\node[anchor=south west,inner sep=0] at (0,0)		{\cropimg{quality/l1b_aug/{\the\numexpr\value{start}+\value{d}*0}}};
		\draw[color=highlight2, line width=1.4pt,rounded corners=false] (114pt,82pt) rectangle (150pt,102pt);
		\draw[color=highlight2, line width=1.4pt,rounded corners=false] (57pt,23pt) rectangle (85pt,50pt);
		\end{tikzpicture}
		\begin{tikzpicture}
		\node[anchor=south west,inner sep=0] at (0,0)		{\cropimg{quality/l1b_aug/{\the\numexpr\value{start}+\value{d}*1}}};
		\draw[color=highlight2, line width=1.4pt,rounded corners=false] (114pt,82pt) rectangle (150pt,102pt);
		\draw[color=highlight2, line width=1.4pt,rounded corners=false] (57pt,23pt) rectangle (85pt,50pt);
		\end{tikzpicture}
		\begin{tikzpicture}
		\node[anchor=south west,inner sep=0] at (0,0)		{\cropimg{quality/l1b_aug/{\the\numexpr\value{start}+\value{d}*2}}};
		\draw[color=highlight2, line width=1.4pt,rounded corners=false] (114pt,82pt) rectangle (150pt,102pt);
		\draw[color=highlight2, line width=1.4pt,rounded corners=false] (57pt,23pt) rectangle (85pt,50pt);
		\end{tikzpicture}
	}
	\caption{Qualitative Results. 
		The images are cropped such that they focus on the interesting regions where differences can be observed.  
		We show three time steps from left to right of RGB input, ground truth, ESPNet with single frame training and ESPNet\_L1b with sequence training (top to bottom).
		Temporally inconsistent predictions by the ESPNet are highlighted using orange boxes.
		The same regions are highlighted by light blue boxes in the predictions of our models to indicate that these regions are predicted consistently.
		Our ESPNet\_L1 model produces high temporally consistent (\unit[98.7]{\%}) results and semantically segments very distant regions, such as the traffic lights and the motorcyclist, correctly.
	}
	\label{fig:quality-suppl}
\end{figure*}
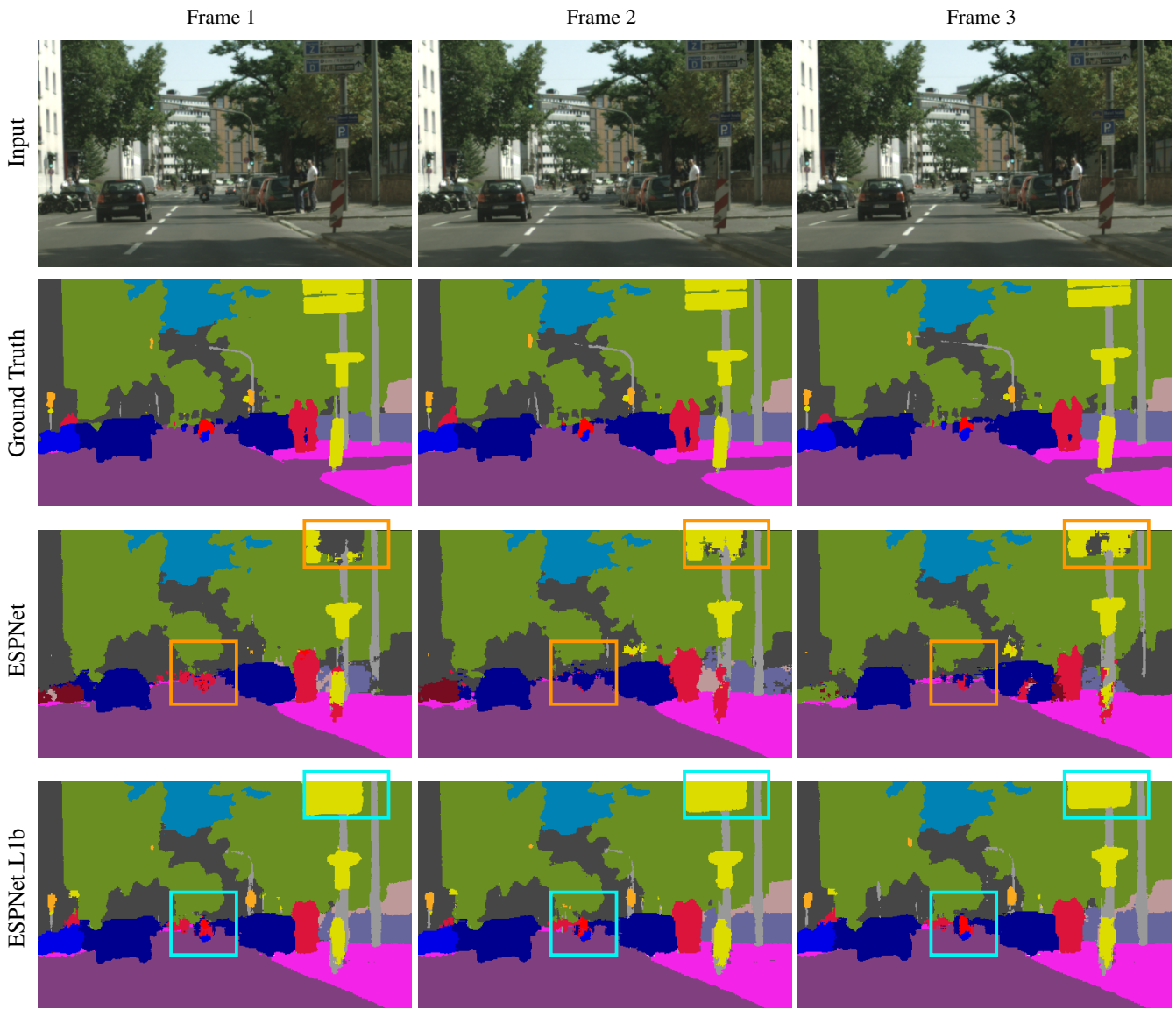

\paragraph{Comparison}
We compare the three convolution types inside the ConvLSTM cell at the output layer of the the ESPNet, \ie architecture ESPNet\_L1a. 
At this position, the ConvLSTM can be interpreted as a post-processing step. 
The input channels of this layer already correspond to the semantic classes. 
Therefore, we can implement the depthwise separable convolution at this position.
We set $C = D = 19$ which is the number of Cityscapes semantic classes and the filter size for this experiment is $P = Q = 3$. 
Consequently, the number of ConvLSTM parameters is
\begin{itemize}
\item \textbf{26,125} for the \textbf{standard convolution},
\item \textbf{1,501} for the \textbf{depthwise separable convolution} and
\item \textbf{79} for the \textbf{depthwise separable convolution with weight sharing}. 
\end{itemize}

The quantitative results are shown in Table \ref{tab:convolution}. We see that the standard convolution performs best in all four metrics. We conclude from the experiment that information from other semantic input channels is needed for a given output channel. Therefore, the standard convolution inside the ConvLSTM is most suitable for our task.

\begin{table}[]
	\setlength\extrarowheight{2pt} 
	\setlength{\tabcolsep}{1.5pt}
	\footnotesize
	\centering
	\begin{tabular}{l|c c c c}
		\hline
		\textbf{Experiment}&\textbf{mIoU}&\textbf{Acc}&\textbf{\Cons}&\textbf{\ConsW}\\ \hline \hline
		Standard Convolution& \textbf{46.5} & \textbf{89.4} & \textbf{97.6} & \textbf{5.4}  \\ \hline
		Depthwise Separable Convolution & 45.2& 89.0  & 97.2  & 5.5 \\ \hline
		Depthw. Sep. Conv. Weight Sharing & 42.5 & 87.7 & 96.9 & 6.5 \\ \hline 
	\end{tabular}
	\caption{ConvLSTM Convolution Types. We compare three different convolution types inside the ConvLST cell using the ESPNet\_L1a architecture. The best results for each metric is highlighted in bold. 
	}\label{tab:convolution}
\end{table}

\section{Qualitative Results}
We qualitatively compare our best model against the ESPNet in Figure \ref{fig:quality-suppl}. 
We observe that the ESPNet shows many temporal inconsistencies and is not able to predict distant objects correctly. 
In contrast, the ESPNet\_L1b produces highly consistent results which are highlighted by the light blue boxes in row four of Figure \ref{fig:quality-suppl}. 
This model is able to predict very small and distant objects correctly although it has the same number of layers than the ESPNet.